\documentclass[10pt,twocolumn,letterpaper]{article}

\usepackage[pagenumbers]{cvpr} 

\usepackage{graphicx}
\usepackage{amsmath}
\usepackage{amssymb}
\usepackage{booktabs}

\usepackage[pagebackref,breaklinks,colorlinks]{hyperref}

\usepackage[capitalize]{cleveref}
\crefname{section}{Sec.}{Secs.}
\Crefname{section}{Section}{Sections}
\Crefname{table}{Table}{Tables}
\crefname{table}{Tab.}{Tabs.}

\def\cvprPaperID{*****} 
\def\confName{CVPR}
\def\confYear{2022}

\usepackage{algorithm}
\usepackage{algorithmic}
\usepackage{multirow}

\usepackage{amsmath,amssymb,bbm,amsthm}
\usepackage{color}
\usepackage{mathtools}

\DeclareMathOperator*{\R}{\mathbb{R}}

\DeclareMathOperator{\prox}{Prox}

\begin{document}

\title{SMOF: Squeezing More Out of Filters Yields Hardware-Friendly CNN Pruning}

\author{Yanli Liu, Bochen Guan, Qinwen Xu, Weiyi Li, and Shuxue Quan\\
OPPO US Research Center\\
InnoPeak Technology\\
Palo Alto, CA, USA\\
{\tt\small \{yanli.liu, bochen.guan, weiyi.li, qinwen.xu, shuxue.quan\}@oppo.com}
}
\maketitle

\begin{abstract}
For many years, the family of convolutional neural networks (CNNs) has been a workhorse in deep learning. Recently, many novel CNN structures have been designed to address increasingly challenging tasks. To make them work efficiently on edge devices, researchers have proposed various structured network pruning strategies to reduce their memory and computational cost. However, most of them only focus on reducing the number of filter channels per layer without considering the redundancy within individual filter channels. In this work, we explore pruning from another dimension, the kernel size. We develop a CNN pruning framework called SMOF, which Squeezes More Out of Filters by reducing both kernel size and the number of filter channels. Notably, SMOF is friendly to standard hardware devices without any customized low-level implementations, and the pruning effort by kernel size reduction does not suffer from the fixed-size width constraint in SIMD units of general-purpose processors. The pruned networks can be deployed effortlessly with significant running time reduction. We also support these claims via extensive experiments on various CNN structures and general-purpose processors for mobile devices.
\end{abstract}

\section{Introduction}

Deep convolutional neural networks (CNNs) have made remarkable breakthroughs on various tasks such as computer vision \cite{Chang_2017_ICCV, Gorji_2018_CVPR}, image processing \cite{Chen_2019_ICCV}, natural language processing \cite{sun2020learning}, and medical imaging applications \cite{ronneberger2015u}. Recently, the rapid development of communication technologies such as Internet-of-Things and 5G \cite{wang2020deep} 
provides a vision of the future where edge devices have a crucial role in providing uninterrupted communications and computations every day. Therefore, there will be an explosive increase in applying deep learning on edge devices to address the new challenges and enable more advisable services. However, successful deep CNN architectures such as ResNet \cite{he2016deep}, DenseNet \cite{huang2017densely}, and EfficientNet \cite{tan2019efficientnet} typically contain hundreds of layers and tons of parameters, which require a large memory footprint and increased computational power. Therefore, there has been a high demand for reducing the size and FLOPs of CNNs with acceptable compromise on their performance.


In view of this challenge, various neural network pruning strategies have been developed. It has been shown that the redundancy in CNNs mostly lies in their convolution filters \cite{zhu2017prune,liu2018rethinking}. In what follows, we classify pruning methods into two main categories: filter channel pruning and filter weight pruning. Overall, their key idea is to obtain high pruning ratios with acceptable loss in performance \cite{9043731}.

Filter channel pruning approaches are developed by removing the filter channels of a CNN according to certain metrics \cite{Meng2020pruning,lin2020hrank}. These methods will change the network structure in a systematic way and only keep basic network blocks to decrease redundancy. Filter channel pruning can achieve high pruning ratios. However, due to the fixed-size width of the SIMD (single instruction, multiple data) unit in general-purpose processors \cite{9022101}, filter channel pruning may not fully utilize the SIMD unit and may not achieve as much of a latency reduction as its FLOPs reduction. 




Filter weight pruning approaches such as \cite{wen2016learning, chen2020orthant} focus on optimizing individual filter weights and achieve a sparse network with acceptable performance degradation. In contrast to channel pruning, weight pruning approaches keep the network structure. Since the sparse weights are often randomly distributed in filters, weight pruning approaches usually need an extra record to locate the sparse weights. However, unstructured sparse convolution does not have an efficient implementation on popular hardware architectures (e.g., ARM and x86) and frameworks (e.g., Tensorflow Lite and TensorRT) \cite{9022101}.  

In summary, filter channel and weight pruning methods are mostly developed on an algorithmic level without considering the structural constraints of current edge devices. Therefore, we ask, if possible, if not more desirable, to develop a deep CNN pruning method that yields high pruning ratios and runtime reduction ratios under hardware limitations. In this work, we propose SMOF to achieve this goal. We summarize our contributions as follows:




\begin{enumerate}
    \item We develop a hybrid pruning method called \textbf{S}queezing \textbf{M}ore \textbf{O}ut of \textbf{F}ilters (SMOF), which considers both filter weight pruning and channel pruning by reducing the kernel size and the number of output channels. 
    
    
    \item SMOF is hardware-friendly. The pruned network can be deployed effortlessly on common hardware devices without any customized low-level implementations and also enjoy significant runtime reduction. Notably, the pruning effort by kernel size reduction does not suffer from the fixed-size width constraint in SIMD units of general-purpose processors.
    
    \item We demonstrate the efficiency of SMOF on various vision tasks. To evaluate its runtime acceleration, SMOF is compared with several state-of-art pruning methods on general-purpose processors for mobile devices.
    
    

\end{enumerate}




\section{Related Work}



\subsection{Filter Channel Pruning}

Filter channel pruning approaches are proposed to remove filter channels within each layer based on different measuring metrics or importance scores. Channel pruning keeps the predefined network structure and decreases redundancy at the layer level. 

Some previous strategies focus on finding better criteria to prune unimportant filter channels. \cite{li2016pruning} shows the feasibility of using the $\ell_2$-norm of filter weights as a measuring metric. \cite{luo2017thinet,lin2020hrank,ijcai2020-363} find that the variance of the feature maps could be another choice for this metric. FPGM \cite{he2019filter} proposes to measure the geometric information of filters for computing importance scores. In addition, \cite{he2017channel} introduce $\ell_1$/$\ell_2$ regularization to modify network channels during training. \cite{liu2017learning} adds Lasso regularization on the batch normalization layer to remove the filter channels with small scaling factors. Recently, there is a trend of applying automated machine learning (AutoML) for automatic network compression \cite{he2018amc,real2019regularized,cai2019once,chin2020towards}. The key idea of these methods is to explore the total space of network filter configurations for the best candidate. On the other hand, Amc \cite{he2018amc} uses reinforcement learning to sample the structure searching space while retaining accuracy, and LeGR \cite{chin2020towards} defines a pair of learnable parameters to adjust the important scores across layers, which leads to a global ranking of filters.

However, these methods only develop ranking and removal mechanisms for filter channels without considering the redundancy within the filters. Moreover, arbitrarily pruned filter channels may not be efficiently implemented on edge devices, where the numbers of output channels should align to the fixed-size width of SIMD units.

\subsection{Filter Weight Pruning}

Filter weight pruning focuses on removing individual filter weights and produce a sparse network. Some methods such as OBProx-SG \cite{chen2020orthant} and SWAT \cite{raihan2020sparse} embed the pruning demand into the network training loss and employs a joint strategy of fine-tuning and optimization to learn a sparse CNN. OICSR \cite{Li_2019_CVPR} applies group Lasso regularization to induce sparsity on filter weights. Another recent work \cite{kusupati2020soft} enhances the sparsity of weights by adding thresholds to each filter. On the other hand, some other methods focus on converting the network to the spectral domain \cite{liu2018frequency,guan2019specnet}, and utilize sparsity of both filters and feature maps to reduce computational cost. However, for the popular edge device hardware architectures such as ARM and x86, these unstructured sparse convolution computations cannot be implemented efficiently. They require an additional record to locate the sparse weights, which takes extra memory and computational resources.

\subsection{Hybrid Filter Pruning}

In view of the filter redundancy in both channels and weights, several pruning methods are proposed to account for both dimensions. Some of them attempt to use AutoML to search a compact network structure.  \cite{cai2019once} and \cite{8897011} propose to search for a compact structure by shrinking a large network from different dimensions, including number of filter channels and kernel size. However, the training is very heavy in both machine time and human labor. On the other hand, SSL \cite{wen2016learning} develops a sparsity learning method to regularize filter weights and channels. In addition, some other approaches add extra components or blocks to help the network learn a sparse structure. For example, SWP \cite{Meng2020pruning} assigns a learnable matrix to each filter weight to learn its desired kernel shape by stripe-wise pruning\footnotemark[1], and remove a filter channel when all of its stripes are pruned. In PCONV \cite{ma2020pconv} and PatDNN \cite{niu2020patdnn}, the kernel shapes can be chosen from a predefined set, and the number of remaining filter channels at each layer should be smaller than some predefined threshold. However, allowing various kernel shapes requires customized implementation on the hardware, as most kernel shapes do not fit in the SIMD unit of processors.  In summary, both regularization methods and kernel shaping methods require a further customized ASIC design \cite{9022101} or special low-level implementation to fully utilize the hardware computational capacity. Otherwise, significant runtime reduction ratios cannot be achieved.

\footnotetext[1]{See Fig. \ref{fig: filter illustration} for the definition of stripe.}

\section{Proposed Method}




In this paper, we propose SMOF for CNN pruning, which \textbf{S}queezes \textbf{M}ore \textbf{O}ut of \textbf{F}ilters by reducing both the kernel size and the number of filter channels at each layer. At a high level, we assign each convolutional layer a learnable matrix called \textit{Filter Skeleton (FS)} to learn the \textit{kernel shape} of filter channels, and another learnable vector called \textit{Filter Mask (FM)} to learn the importance of individual filter channels. With the same square kernel shape across filter channels at each layer, the pruned network can be deployed effortlessly on popular hardware devices without customized low-level implementations. Notably, the pruning effort by kernel size reduction is not canceled by the fixed-size width of SIMD units in processors.



\begin{figure}
    \centering
    \includegraphics[width = 0.45\textwidth]{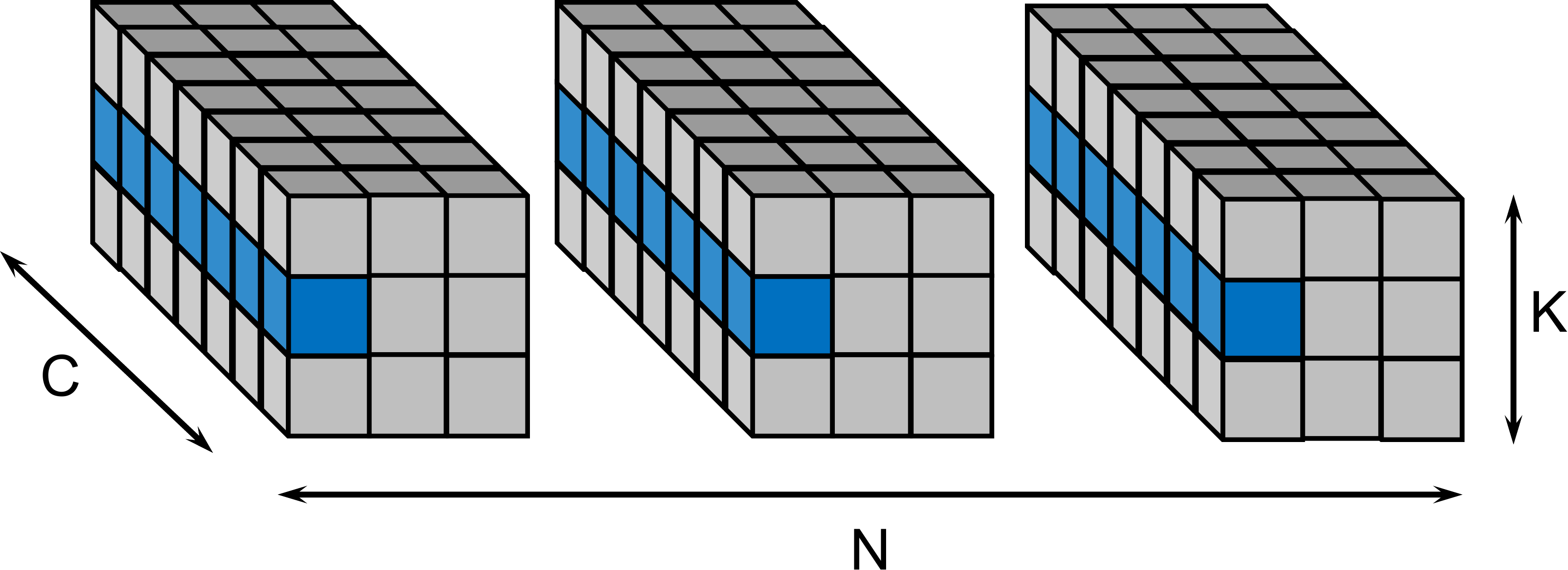}
    \caption{\textbf{Stripes of a convolution layer.} Each filter channel has a square kernel shape with kernel size $K$, and has $K^2$ stripes. The stripes in blue are at the same position across filter channels, they will be pruned if their corresponding FS value is small.}
    \label{fig: filter illustration}
\end{figure}

More specifically, let the weight of a convolutional layer be $W\in\R^{N\times C\times K\times K}$, where $K$ is the \textit{kernel size}, and $N, C$ are the number of output and input channels, respectively. We say that each filter channel consists of $K^2$ \textit{stripes} of the form $\R^{C\times 1}$ (see Fig. \ref{fig: filter illustration} for an illustration). In order to ensure an efficient implementation of the pruned network on hardware devices, we propose SMOF with a focus on reducing $K$ and $N$ in a learnable and coordinated way, see Algorithm \ref{alg: SMOF} for details.




\subsection{Filter Skeleton (FS) for Kernel Size Reduction}

\begin{algorithm}
\caption{SMOF: Squeezing More Out of Filters yields hardware-friendly CNN pruning }
\label{alg: SMOF}
    \begin{algorithmic}[1]
    \STATE{\textbf{Input}: CNN with initial convolution weights $W$, penalty parameters $\alpha$, $\beta$ for Filter Skeleton (FS) and Filter Mask (FM), pruning parameters $\rho_{\text{FS}}$, $\delta_{\text{FM}}$ for FS and FM. Total Number of iterations $K$;}
    \STATE{\textbf{Output}: pruned \& trained CNN;}
    \STATE{Initialize the elements of FS and FM as 1;}
        \FOR{$k = 1,2,...,K$}
        \STATE{Update FS of each layer in the ``peeling" fashion in Fig. \ref{fig: FS pruning}, with ratio $\rho_{\text{FS}}$;}
        \STATE{Update FM of each layer by the procedures in Fig. \ref{fig: FM pruning}, with threshold $\delta_{\text{FM}}$;}
        \STATE{Apply one iteration of SGD to $W$ and non-zero elements of FM, with the full training loss in \eqref{equ: loss with FS and FM};}
        \STATE{Apply one iteration of proximal gradient descent given by \eqref{equ: update of FS} to nonzero elements of FS;}
        \ENDFOR
    \STATE{$W \leftarrow W\odot \text{FS}\odot \text{FM}$;}
    \STATE{Reduce kernel sizes and number of filter channels at each layer by procedures described in Figs. \ref{fig: FS pruning} and \ref{fig: FM pruning};}
    \STATE{Adjust zero padding at each layer based on its kernel size reduction;}
    \end{algorithmic}
\end{algorithm}

\begin{figure}
    \centering
    \includegraphics[width = 0.26\textwidth]{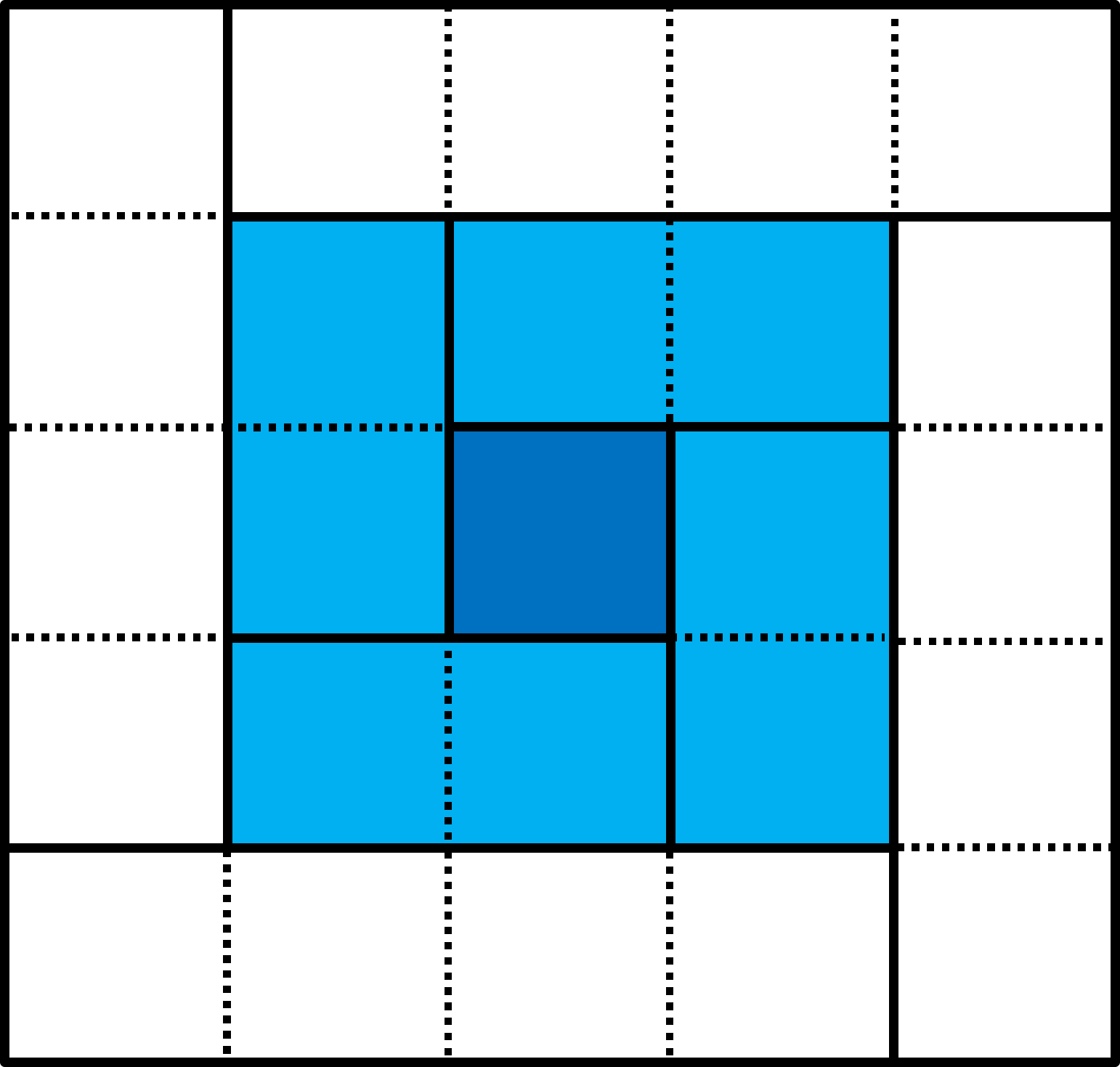}
    \caption{\textbf{Kernel size reduction by FS Pruning.} This FS of size $5\times 5$ has 2 slices. For each slice, its $4$ edges are in solid lines. For the outermost slice with $i=1$ (in white), its FS elements have been pruned to $0$ at some previous iteration. The current kernel size is $3$. We now proceed to its inner slice with $i=2$ (in light blue). During training, if the sum of the absolute values of its FS elements is smaller than $\rho_{\text{FS}}\cdot 4(K^l + 1 -2i)$, these elements will be pruned to $0$, and the kernel size is further reduced from $3$ to $1$. The central element (in dark blue) will never be pruned in SMOF. To prepare for inference, we multiply FS to the weight $W$ and prune the outer stripes of $W$ with zero FS value.}
    \label{fig: FS pruning}
\end{figure}

Let $W^l\in\R^{N^l\times C^l\times K^l\times K^l}$ be the weight of $l$th convolution layer. Inspired by \cite{Meng2020pruning}, we assign a learnable parameter called Filter Skeleton $\text{FS}^l\in\R^{K^l\times K^l}$ to $W^l$ to reduce the kernel size $K^l$. Each element in FS reflects the importance the corresponding stripe in all of the $N^l$ filter channels of $W^l$, and is initialized as $1$. Mathematically, the training loss with FS can be written as:
\begin{align}
\label{equ: loss with FS}
\begin{split}
&\text{Training loss with FS:}\\
&\sum_{(x,y)\in D} \text{loss}(f(x, W\odot FS), y) + \sum_l\sum_{i = 1}^{K^l//2} \alpha^l_i \|\text{FS}_i^l\|_{g},
\end{split}
\end{align}
where $K^l$ is the kernel size of the $l$th convolutional layer, and $\text{FS}^l_i$ denotes the $i$th \textit{slice} of $\text{FS}^l$ (see Fig. \ref{fig: FS pruning}), and $\odot$ denotes the dot product. The penalty term in \eqref{equ: loss with FS} promotes kernel size reduction. Specifically, the $\|\cdot\|_g$-norm induces group sparsity on the 4 edges $\text{FS}^l_i$ ($i=1,2,3,4$).
\begin{align}
\label{equ: group sparsity}
\begin{split}
\|\text{FS}_i^l\|_{g} &= \sum_{j=1}^4 \|\text{FS}_{i,j}^l\|_{2}.
\end{split}
\end{align}
the $\|\text{FS}_i^l\|_{g}$ term induces structured sparsity on the 4 edges $\text{FS}_{i,j}^l$ ($j=1,2,3,4$) of $\text{FS}^l_i$, which is inspired from Group Lasso \cite{simon2013sparse,roth2008group}, This type of penalty term has also been applied in other network pruning methods \cite{liu2015sparse,wen2016learning}.


Kernel size reduction is done in a ``peeling" fashion during training, as illustrated in Fig. \ref{fig: FS pruning}. More specifically, we set a certain ratio $\rho_{\text{FS}}>0$ during training. Starting from $i=1$, we consider the $4$ edges on the $i$th slice $\text{FS}_i^l$. If at some iteration during training, the sum of the absolute values of elements on $\text{FS}^l$ on these edges are smaller than\footnotemark[1] $\rho_{\text{FS}}\cdot 4(K^l + 1 - 2i)$, we will prune these elements to $0$. After $\text{FS}^l$ is multiplied to $W^l$, the stripes with positions that correspond to zero values in $\text{FS}^l$ will also be zero, and the kernel size of $W^l$ is reduced by $2$, We repeat this process on the remaining $\text{FS}^l$ elements (i.e., $i\leftarrow i+1$) to further reduce kernel size. Note that the most central element will not be pruned to 0. To keep current progress, the $\text{FS}^l$ elements that are pruned to 0 will not be updated at later iterations. 


\footnotetext[1]{$4(K^l + 1 - 2i)$ is the initial sum of the FS elements on 4 edges.}

In \eqref{equ: loss with FS}, the coefficients $\alpha^l_i$ are of the form: \begin{align}
\label{equ: alpha definition}
\alpha^l_i = (K^l//2 + 1 - i)\alpha \quad \text{for some $\alpha>0.$}
\end{align} 
This gives a stronger penalty to the outer slices of $\text{FS}^l$, and encourages pruning to start from outer slices.




\subsection{Filter Mask (FM) for Filter Number Reduction}

To further improve pruning ratios, we also reduce the number of filter channels at each layer alongside kernel size reduction. For that purpose, we assign a learnable vector called Filter Mask $\text{FM}^l\in\R^{N^l\times 1}$ to the weight $W^l\in\R^{N^l\times C^l\times K^l\times K^l}$. $\text{FM}^l$ is multiplied to $W^l\odot\text{FS}^l$ during training, which learns the importance of the $N^l$ output channels. The elements of $\text{FM}^l$ are initialized as $1$.


Together with FS, we now have the full training loss:
\begin{align}
\label{equ: loss with FS and FM}
\begin{split}
&\text{SMOF Training loss with FS and FM:}\\
&\sum_{(x,y)\in D} \text{loss}(f(x, W\odot FS\odot FM), y) \\
& + \sum_l\sum_{i = 1}^{K^l//2} \alpha^l_i \|\text{FS}_i^l\|_{g} + \sum_{l}\beta \|\text{FM}^l\|_1,
\end{split}
\end{align}
where $\alpha^l_i$ is given by \eqref{equ: alpha definition}, and the last term applies $\ell_1$-norm penalty to induce sparsity on $\text{FM}^l$. During training, we set a threshold $\delta_{\text{FM}}>0$, and prune the $\text{FM}^l$ elements with absolute values smaller than $\delta_\text{FM}$ to $0$. After multiplying $\text{FM}^l$ to $W^l\odot\text{FS}^l$, any filter channel with a 0 in $\text{FM}^l$ is also 0.

For CNNs with special structures, pruning channels at different layers arbitrarily may destroy these structures and cause significant performance degradation. For example, the shortcut connection is a key feature in ResNet \cite{he2016deep}. It leads to the superior performance of ResNet over earlier network designs such as GoogleNet \cite{szegedy2015going} and VGGNet \cite{Simonyan15}. In a simple form, the shortcut connection is given by
\begin{align}
\label{equ: shortcut}
y = \mathcal{F}(x, \{W^l\}) + x.
\end{align}
When $W^l$ is pruned by $\text{FM}^l$, it will have fewer output channels, and a dimension mismatch between $\mathcal{F}(x, \{W^l\})$ and $x$ may occur. 

\begin{figure}[ht]
    \centering
    \includegraphics[width = 0.45\textwidth]{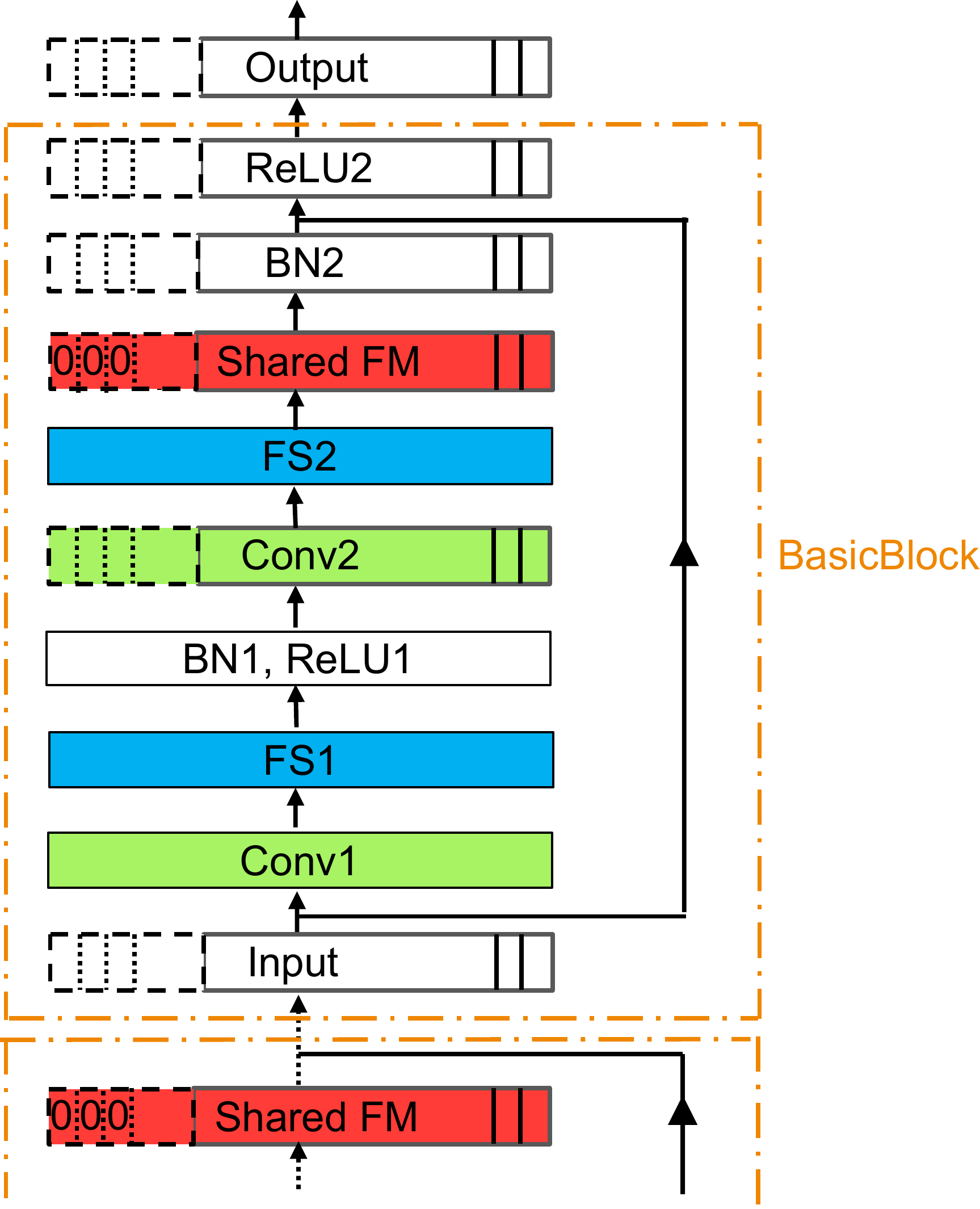}
    \caption{\textbf{Filter channel reduction by FM pruning.} For simplicity, we focus on the case of the simple form of shortcut connection \eqref{equ: shortcut} in BasicBlock. FM pruning applies to other structures with shortcut connections as well. The orange blocks denote BasicBlocks with the same structure, in which the shared FM is applied after the FS2 block. The dashed part in shared FM has been pruned to 0. As a result, the corresponding output channels in Conv2 are also pruned to 0. Since the same shared FM is also applied in the previous BasicBlock, the shortcut connection can be applied safely. To prepare for inference, we prune the dashed output channels in the network.}
    \label{fig: FM pruning}
\end{figure}

To resolve this, we apply a shared FM to all the layers that must have the same output dimension, and prune their output channels at the same time (see Fig. \ref{fig: FM pruning}). In other words, the shared FM reflects the importance of their output channels altogether. Take ResNet18 as an example, $4$ shared FMs are applied for $4$ different BasicBlocks. 

\begin{table*}
\centering
\begin{tabular}{c|cccccc}
\hline
Method   & \begin{tabular}[c]{@{}c@{}} Params\\ \% $\downarrow$ \end{tabular} & \begin{tabular}[c]{@{}c@{}} FLOPs\\ \% $\downarrow$ \end{tabular} & \begin{tabular}[c]{@{}c@{}} CPU\\ \% $\downarrow$ \end{tabular} & \begin{tabular}[c]{@{}c@{}} GPU\\ \% $\downarrow$ \end{tabular} & \begin{tabular}[c]{@{}c@{}} DSP\\ \% $\downarrow$ \end{tabular} &  \begin{tabular}[c]{@{}c@{}} Accuracy\\ \%  \end{tabular} \\ \hline\hline
\begin{tabular}[c]{@{}c@{}}SMOF: $\alpha = 1e-4$,  $\rho_{\text{FS}} = 0.425$, \\ $\beta = 1e-3$,  $\delta_{\text{FM}} = 0.2$, $r = 1/2$\end{tabular}   &      48.28                    &       60.50     & 3.57    & 45.61     & 21.30    & 93.18 $\rightarrow$ 93.57                    \\ \hline
SWP \cite{Meng2020pruning} &   77.7     &    75.6      &  -        &  -   & -    & 93.10 $\rightarrow$ 92.98      \\ \hline
HRank \cite{lin2020hrank} &   42.4     &    50.0      &   -1.46       &  26.32   & 12.43      &  93.26 $\rightarrow$ 93.17     \\ \hline
Global Ranking \cite{chin2020towards} &    -    & 53        & 0.32     & 29.82     & 13.01      &  93.9 $\rightarrow$ 93.7        \\ \hline
\end{tabular}
\caption{\textbf{ResNet56 pruning with CIFAR-10.} The baseline ResNet56 model has a CPU runtime of 616(ms), a GPU runtime of 5.7(ms), and a DSP runtime of 0.845(ms). All DSP runtime are averaged over 20 runs. SMOF achieves the most runtime reduction with an accuracy gain of $0.39\%$. More results and comparisons can be found in Sec. \ref{sec: ablation}.}
\label{table: ResNet56 results}
\end{table*}

\subsection{Training and Inference}



In order to perform training with the loss function \eqref{equ: loss with FS and FM}, we apply stochastic gradient descent (SGD) on $\{W^l\}$ and $\{\text{FM}^l\}$. To deal with the nondifferentiable penalty term on $\{\text{FS}^l\}$, we apply proximal gradient descent \cite{bauschke2011convex}. More specifically, the update on $\text{FS}^l$ is given by 
\begin{align}
\label{equ: update of FS}
\text{FS}^{l,+}_{i,j} &= \prox_{\eta\alpha_i \|\cdot\|_2}(\text{FS}^l_{i,j} - \eta \alpha_i \tilde{g}^{l}_{i,j}),
\end{align}
where $\eta$ is the learning rate, $i = 1,2,...,K^l//2$, and $ j = 1,2,3,4.$ $\tilde{g}^{l}_{i,j}$ is a stochastic gradient of the first term in \eqref{equ: loss with FS and FM} w.r.t. $\text{FS}^l_{i,j}$.
The proximal operator $\prox_{\eta\alpha_i \|\cdot\|_2}$ has a closed form expression, which is cheap and easy to implement:
\begin{align}
\label{equ: prox on group sparsity closed form}
\prox_{\eta\alpha_i \|\cdot\|_2}(x) = \frac{x}{\|x\|_2}\max\{0, \|x\|_2 - \eta\alpha_i\}.
\end{align}



After training is finished, we multiply $\text{FS}^l$ and $\text{FM}^l$ to $W^l$, and prune the outer stripes of $W^l$ with zero value in $\text{FS}^l$, as well as output channels with zero value in $\text{FM}^l$ (see Figs. \ref{fig: FS pruning} and \ref{fig: FM pruning}). 
We also need to adjust the zero padding. For example, if the kernel size of $W^l$ is reduced by $2$, its zero padding should be decreased by $1$. After this step, the pruned model is ready for inference.



\section{Experiments}
\label{sec: experiments}

To validate our proposed approach, we conduct extensive experiments on various deep CNN architectures including VGG \cite{Simonyan15}, ResNet \cite{he2016deep} for image classification, and UNet \cite{ronneberger2015u} for image denoising.

\noindent\textbf{Hardware Settings} We use SNPE (Snapdragon Neural Processing Engine SDK) on Qualcomm Snapdragon SM8250 as our mobile platform. In the following, we test SMOF and other CNN pruning methods on three different general-purpose processors for mobile devices: CPU, GPU, and digital signal processor (DSP). The CPU is a Kryo 585 CPU (ARM V8 Cortex based architecture CPU at 1.8 - 2.84GHz). Our GPU is Adreno 650 GPU with 1.2T FLOPS. Our DSP is Hexagon 698 V66 processor, which is an efficient low power general processor with 15TOPS. These are typical hardware accelerators for deploying CNNs on mobile devices. It is worth noting that the implementation on  CPU, GPU, and DSP will try to pad the output channels of Conv layers to be aligned to fixed-size width to fully utilize the SIMD units. For example, the Qualcomm Hexagon DSP V66 processor has a SIMD vector of 1024 bits. All models are trained on a 8 NVIDIA A100 GPUs server using PyTorch.




\subsection{ResNet56 and VGG16 on CIFAR-10}
\label{subsec: ResNet56 and VGG16 test}


\noindent\textbf{Models and Dataset} We test on the CIFAR-10 dataset \cite{krizhevsky2009learning}, which contains 50K training images and 10K test images for 10 classes. For the training set, we apply standard data augmentation procedures, including random crop and random horizontal flip. For this dataset, we apply two popular CNNs: VGG16 \cite{Simonyan15} and ResNet56 \cite{he2016deep}.  

\noindent\textbf{Parameter Settings} For ResNet56 and VGG16, we start pruning from pretrained models. We train for $180$ epochs, where the initial learning rate is $0.1$ and is divided by $10$ at the $90$th and $135$th epoch. The training algorithm is SGD with a momentum of $0.9$, a batchsize of $128$, and a weight decay of $1e-4$. During the first 90 epochs, we apply parameter settings listed in Tables \ref{table: ResNet56 results} and \ref{table: VGG16 results}. After the 90th epoch, we set $\alpha = \beta = \rho_{\text{FS}} = \delta_{\text{FM}} = 0$, which stops the pruning process and fixes the obtained network structure.  

For ResNet56, we also set a certain percentage (denoted by $r$) of FM values to be learnable at each convolution layer, while the rest are always $1$. This avoids pruning of all of the channels in one layer and therefore makes training easier\footnotemark[1].

\footnotetext[1]{If we set $r = 1$, then for certain layers of ResNet56, the elements of FM are very similar during training, and all the channels are pruned nearly at the same time. By fixing some FM elements to be $1$, we keep the corresponding channels and allow channel pruning for the rest.}

\begin{center}
\begin{table}
\scalebox{0.95}{
\begin{tabular}{c|ccc}
\hline
Method   & \begin{tabular}[c]{@{}c@{}}Params\\ \% $\downarrow$ \end{tabular}& \begin{tabular}[c]{@{}c@{}} FLOPs \\ \% $\downarrow$ \end{tabular}& \begin{tabular}[c]{@{}c@{}} Accuracy \\ \% \end{tabular}\\ \hline\hline
\begin{tabular}[c]{@{}c@{}}SMOF: $r = 1$ \\ $\alpha = 1e-6$,  $\rho_{\text{FS}} = 0.31$, \\ $\beta = 1e-8$,  $\delta_{\text{FM}} = 0.02$\end{tabular} &                80.8         &        50.1               &        93.88     \\ \hline
\begin{tabular}[c]{@{}c@{}}SMOF: $r = 1$ \\ $\alpha = 1e-6$,  $\rho_{\text{FS}} = 0.31$, \\ $\beta = 1e-8$,  $\delta_{\text{FM}} = 0.025$\end{tabular} &  80.9            &         58.0          &       93.50            \\ \hline
HRank \cite{lin2020hrank} &   82.9     &      53.5          &  93.43  \\ \hline
GAL-0.1 \cite{lin2019towards} &  82.2      &  45.2              &   93.42   \\ \hline
Zhao \textit{et al.} \cite{zhao2019variational} & 73.3       &  39.1              &   93.18   \\ \hline
\end{tabular}}
\caption{\textbf{VGG16 pruning with CIFAR-10.} The baseline of SMOF has an accuracy of 93.95\%. The baseline of HRank \cite{lin2020hrank} and GAL-0.1 \cite{lin2019towards} has an accuracy of 93.96\%. The baseline of Zhao \textit{et al.} has an accuracy of 93.25\%.}
\label{table: VGG16 results}
\end{table}
\end{center}

\noindent\textbf{Discussion of ResNet56 Results} In Table \ref{table: ResNet56 results}, we present a pruning ratio and runtime comparison between our SMOF and other pruning strategies. Our baseline model has an accuracy of $93.18\%$. SMOF achieves a higher pruning ratio than HRank \cite{lin2020hrank} and Global Ranking \cite{chin2020towards}, while enjoying a $0.39\%$ accuracy gain. SMOF also saves much more GPU and DSP runtime than HRank \cite{lin2020hrank} and Global Ranking \cite{chin2020towards}. We speculate that this is because these two strategies only focus on filter channel pruning, while for processors with longer SIMD (e.g., 1024 bit for the DSP processor we used for test), the aligning processing explained at the beginning of Sec. \ref{sec: experiments} cancels some of their pruning efforts. Moreover, the data padding/cropping itself costs extra latency. In contrast, most of SMOF's pruning efforts lie in kernel size reduction, which is not constrained by this hardware limitation.

Compared with SMOF, SWP \cite{Meng2020pruning} has higher pruning ratios as it achieves finer granularity by allowing pruning of individual stripes at each layer. However, the pruned channels typically have irregular kernel shapes, which also vary a lot among channels at the same layer. As a result, customized low-level implementations are required to achieve runtime reduction. 




\noindent\textbf{Discussion of VGG16 Results} We summarize the results in Table \ref{table: VGG16 results}, where runtime comparison is not included\footnotemark[1]. When compared with HRank \cite{lin2020hrank} and GAL-0.1 \cite{lin2019towards}, SMOF in the second row has a higher FLOPs reduction ratio as well as a smaller accuracy drop. SMOF in the first row achieves a similar accuracy drop as \cite{zhao2019variational} and more FLOPs reduction.



\footnotetext[1]{Runtime comparison is not included here since pruned models of HRank and GAL-0.1 in Table \ref{table: VGG16 results} are not available.}

\begin{table*}
\centering
\begin{tabular}{c|cccccc}
\hline
Method   & \begin{tabular}[c]{@{}c@{}} Params\\ \% $\downarrow$ \end{tabular} & \begin{tabular}[c]{@{}c@{}} FLOPs\\ \% $\downarrow$ \end{tabular} & \begin{tabular}[c]{@{}c@{}} CPU\\ \% $\downarrow$ \end{tabular} & \begin{tabular}[c]{@{}c@{}} GPU\\ \% $\downarrow$ \end{tabular} & \begin{tabular}[c]{@{}c@{}} DSP\\ \% $\downarrow$ \end{tabular} &  \begin{tabular}[c]{@{}c@{}} Accuracy\\ \%  \end{tabular} \\ \hline\hline
SMOF-1   &      24.96                    &       31.02     & 18.64    & 28.81     & 29.44    & 69.76 $\rightarrow$ 69.21                    \\ \hline
SMOF-2   &      25.21                    &       36.60     & 19.08    & 32.92     & 33.47    & 69.76 $\rightarrow$ 68.55                    \\ \hline
SWP \cite{Meng2020pruning} &   -     &    54.48      &  -        &  -   & -    & 69.76 $\rightarrow$ 69.59      \\ \hline
DMCP \cite{guo2020dmcp} &   -     &  43.54             &  22.08  & 34.57    & 11.29 &    70.1 $\rightarrow$ 69.2 \\ \hline
ResRep$^*$ \cite{ding2020lossless} &  -   & 40.61   &  16.96   &  33.33    & 19.35 & 69.76 $\rightarrow$ 69.12 \\ \hline
FPGM \cite{he2019filter} &   -     &  36.48       &  17.67    &  36.63    & 15.32      &  69.76 $\rightarrow$  68.54       \\ \hline
\end{tabular}
\caption{\textbf{ResNet18 pruning with ImageNet.} The pruned models SMOF-1 and SMOF-2 are obtained with parameters and procedures described in Sec. \ref{subsec: ResNet18 test}. The baseline ResNet18 model has a CPU runtime of 566(ms), a GPU runtime of 24.3(ms), and a DSP runtime of 4.96(ms). $*$ denotes our independent test. All DSP runtime are averaged over 20 runs. }
\label{table: ResNet18 results}
\end{table*}

\subsection{ResNet18 on ImageNet}
\label{subsec: ResNet18 test}

\noindent\textbf{Models and Dataset} We evaluate SMOF to ResNet18 \cite{he2016deep} on the popular ImageNet dataset \cite{deng2009imagenet}, which contains 1.28 million training images and 50K test images for 1000 classes. 


\noindent\textbf{Parameter Settings} Our training settings are similar to the official PyTorch implementation\footnotemark[2]. Specifically, we apply SGD with momentum $= 0.9$, an initial learning rate of $0.1$, a batchsize of $256$, and a weight decay factor of $1e-4$. 

\footnotetext[2]{\url{https://github.com/pytorch/examples/tree/master/imagenet}}


Throughout this test, we keep all the elements in FilterMasks to be $1$ (i.e., $r = 0$). First, we set $\alpha = 1e-6$ and $\rho_{\text{FS}} = 0$ for $10$ epochs. Then, we either 1) set $\alpha = 2e-4$ and $\rho_{\text{FS}} = 0.45$ for 20 epochs, or 2) set $\alpha = 1e-6$ and $\rho_{\text{FS}} = 0.625$ for 20 epochs. We call the obtained models SMOF-1 and SMOF-2, respectively. Next, we stop pruning and fix their network structures by setting $\alpha = \rho_{\text{FS}} = \beta = \delta_{\text{FM}} = 0$, and apply learning rates $1e-1, 1e-2, 1e-3, 1e-4$ for 20 epochs, respectively.


\noindent\textbf{Discussion} In Tables \ref{table: ResNet18 results} and \ref{table: ResNet18 results}, we present a pruning ratio and runtime comparison between our SMOF and other pruning strategies. The SMOF-1 model has smaller pruning ratios and a higher accuracy compared with SMOF-2. Similar as in the ResNet56 test, SWP \cite{Meng2020pruning} achieves the best pruning ratios by allowing pruning of individual stripes, but customized low-level implementations are required to achieve runtime reduction on processors. DMCP \cite{guo2020dmcp}, ResRep \cite{ding2020lossless}, and \cite{he2019filter} are channel pruning methods, they have a similar accuracy compared to SMOF-1 or SMOF-2. Although they have higher FLOPs reduction ratios, their CPU and GPU runtime are similar to SMOF, and their DSP runtime are worse. This is because the aligned processing mentioned at the beginning of Sec. \ref{sec: experiments} cancels some of their channel pruning efforts, especially on the DSP processor with a longer SIMD unit of 1024 bits. Moreover, the data padding/cropping itself costs extra latency.


We provide structure of pruned ResNet18 in appendix.

\subsection{U-Net for Image Denoising}

\noindent\textbf{Models and Dataset} In this section, we evaluate SMOF on U-Net for image denoising, and compare with five state-of-the-art pruning approaches (DHP \cite{10.1007/978-3-030-58598-3_36}, Factor-SIC2 \cite{wang2017factorized}, Group \cite{peng2018extreme}, LeGR \cite{chin2020towards}, and HRank \cite{lin2020hrank}). We train on the DIV2K \cite{Agustsson_2017_CVPR_Workshops} dataset and evaluate on the BSD68 dataset.  

\noindent\textbf{Parameter Settings} To ensure a fair comparison, we apply the same network hyperparameters as in the original network training strategy \cite{ronneberger2015u}. We use SGD with momentum$=0.9$, fixed learning rate of $1e-4$, a batchsize of 256, and a weight decay factor of $5e-5$. We also apply the same procedures for data preprocessing, filter weights initialization, and the same noise level of images (70). To compare these model compression methods, we measure their performance in six metrics, including the number of parameters, FLOPs, Peak Signal-to-Noise Ratio (PSNR), and runtime on CPU, GPU, and DSP. 

\begin{table*}[t]
\centering
\scalebox{1.0}{
\begin{tabular}{c|c|ccc|ccc} 
\hline
Network               & Method       & \begin{tabular}[c]{@{}c@{}}Params\\ \% $\downarrow$ \end{tabular} & \begin{tabular}[c]{@{}c@{}}FLOPs\\ \% $\downarrow$ \end{tabular} & \begin{tabular}[c]{@{}c@{}}PSNR\\  $\downarrow$ \end{tabular} & \begin{tabular}[c]{@{}c@{}}CPU\\ (ms) \end{tabular} & \begin{tabular}[c]{@{}c@{}}GPU\\ (ms) \end{tabular}       & \begin{tabular}[c]{@{}c@{}}DSP\\ (ms) \end{tabular}  \\ 
\hline
\multirow{7}{*}{UNet} & DHP   \cite{10.1007/978-3-030-58598-3_36}     & 58.25      & 58.06     & 0.12      & 163.9    & 9.85             & 0.71      \\
                      & Factor-SIC2 \cite{wang2017factorized} & 67.65      & 64.22     & 0.23      & 828.7    & - & 31.80     \\
                      & Group  \cite{peng2018extreme}   & 73.45      & 56.30     & 0.11      & 294.4    & - & 19.80     \\
                      & LeGR$^*$  \cite{chin2020towards}  & 53.09      & 58.65     & 0.13      & 393.7    & 29.20            & 1.44      \\
                      & Hrank$^*$  \cite{lin2020hrank}     & 56.96      & 57.77     & 0.15      & 157.3    & - & 0.85      \\ 
\cline{2-8}
                      & SMOF-1    & 74.74      & 52.79     & 0.11      &  134.7   & 4.60             &   0.54     \\
                      & SMOF-2      & 58.89      & 44.28     & 0.05      & 152.9    & 8.47              & 0.56     \\
\hline
\end{tabular}}
\caption{\textbf{Comparison of several model pruning methods for image denoising.} The noise level of training and testing images is 70. All the methods are tested on BSD68 dataset and FLOPs is reported for a $128\times128$ image. SMOF achieves significant runtime reduction on different general-purpose processors of mobile devices. $*$ denotes our implementation.}
\label{Tab:denoising}
\end{table*}

\begin{figure*}
    \centering
    \includegraphics[width = 0.98\textwidth]{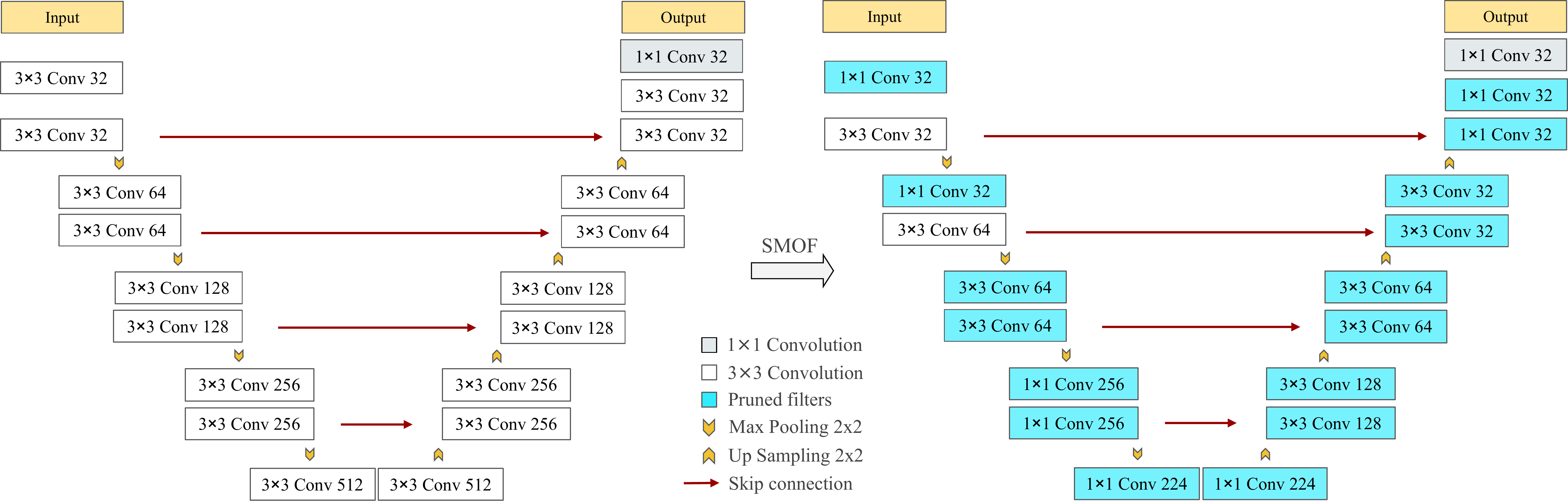}
    \caption{\textbf{The baseline U-Net and pruned U-Net structure (SMOF-1) for image denoising.} The pruned filters (in blue) have either smaller kernel sizes, or less output channels, or enjoy both.} 
    \label{fig:UNet_structure}
\end{figure*}

\noindent\textbf{Discussion} We evaluate two models pruned by SMOF for different purposes:  one for computational efficiency (SMOF-1) and the other one for high PSNR (SMOF-2). We report pruned network structure in Figure \ref{fig:UNet_structure} and summarize the test results in Table \ref{Tab:denoising}. The procedures for obtaining these pruned models can be found in the appendix.

Compared with LeGR \cite{chin2020towards} and HRank \cite{lin2020hrank}, SMOF-1 saves more runtime and has a similar PSNR. Besides, we notice that SMOF-1 achieves the similar PSNR and complexity reduction as Group \cite{peng2018extreme}, but with less run-time on CPU and DSP. This might due to the inefficient implementations of group convolution on mobile device accelerators \cite{10.1007/978-3-030-58598-3_36}. We also notice that, although DHP \cite{10.1007/978-3-030-58598-3_36} and Factor-SIC \cite{wang2017factorized} achieve greater reduction of FLOPs, SMOF-1 performs more efficiently on the processors. 



SMOF-2 obtains an excellent PSNR of 25.19, which is very close to 25.24 of the baseline model. Notably, SMOF-2 has similar or less pruning ratios compared with other approaches, but it performs more efficiently on the processors. As described at the beginning of Sec. \ref{sec: experiments}, this is due to the fixed-size width constraint in the processors.

\begin{figure*}[t]
    \centering
    \includegraphics[scale = 0.45]{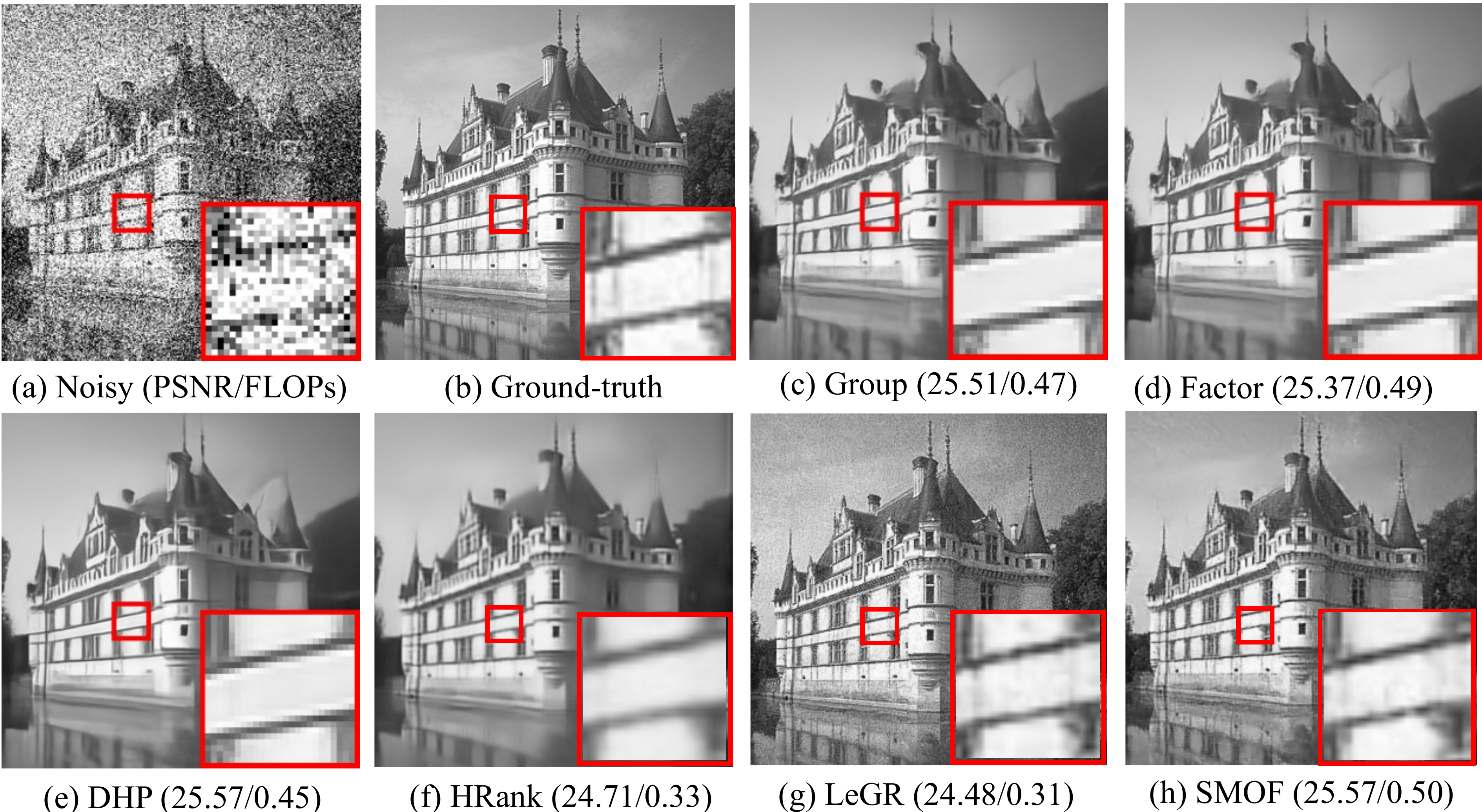}
    \caption{\textbf{Denoising on an image from BSD68 (noise level 50).} We report PSNR and FLOPs measured on this image.}
    \label{fig:denoising}
\end{figure*}

Fig. \ref{fig:denoising} presents the output of SMOF and other algorithms. In terms of the visual quality, the output of SMOF-1 is close to the ground truth. Besides, SMOF remains detailed patterns, which demonstrates the capacity of SMOF to handle Gaussian noise with large noise levels.

\section{Ablation Study}
\label{sec: ablation}


In this section, we demonstrate the advantage of adaptive group sparsity term \eqref{equ: group sparsity} over the usual $\ell_1$-norm. Trade-off curves of SMOF and HRank\cite{lin2020hrank} are also presented.

\begin{figure}[H]
    \centering
    \includegraphics[width=0.48\textwidth]{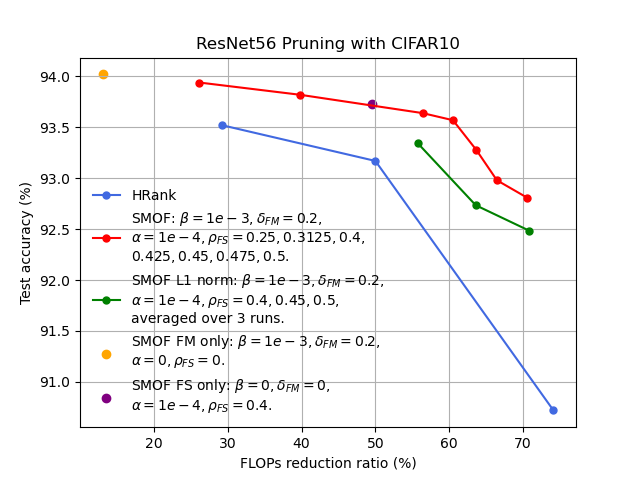}
    \caption{Ablation Study on ResNet56.}
    \label{fig: ablation}
\end{figure}

From Fig. \ref{fig: ablation} we can see that SMOF has a better trade-off curve than HRank. Compared with using the plain $\ell_1$-norm penalty as in \cite{Meng2020pruning}, adaptive group sparsity term \eqref{equ: group sparsity} have a clear advantage as it applies stronger group penalty on the outer edges, which encourages pruning in the ``peeling" fashion as described in Fig. \ref{fig: FS pruning}. On the other hand, the $\ell_1$-norm penalty applies uniform penalty on all the stripes, which is unnecessary for stripes on the inner edges. Finally, when FM(Filter Mask) pruning is disabled, SMOF applies less pruning with the same FS parameters, the same happens for SMOF without FS pruning.

\section{Conclusions and Future Work}


In this work, we propose a hardware-friendly CNN pruning framework called SMOF. The key idea is to Squeeze More Out of Filters by reducing both the kernel size and filter channels at each layer. SMOF learns the importance of stripes and channels. It prunes the unimportant stripes and channels in a coordinated way, such that the kernel size and number of channels can be reduced while preserving the network structure. The efficiency of SMOF is demonstrated on popular CNNs and general-purpose processors without any customized low-level implementations.


There are still open problems to be addressed. For instance, can we allow additional kernel shapes other than just squares? A direct extension to this would be rectangle kernels with different zero padding applied along the x and y directions. In addition, the performance of SMOF enables us to conclude that kernel size reduction is a direction worth pursuing. Therefore, is it possible to combine it with other channel pruning strategies such as \cite{chin2020towards} and \cite{ding2020lossless}? We believe that this would lead to even higher pruning ratios and runtime reduction.

{\small
\bibliographystyle{ieee_fullname}
\bibliography{references}

\begin{thebibliography}{10}\itemsep=-1pt

\bibitem{Agustsson_2017_CVPR_Workshops}
Eirikur Agustsson and Radu Timofte.
\newblock Ntire 2017 challenge on single image super-resolution: Dataset and
  study.
\newblock In {\em The IEEE Conference on Computer Vision and Pattern
  Recognition (CVPR) Workshops}, July 2017.

\bibitem{bauschke2011convex}
Heinz~H Bauschke, Patrick~L Combettes, et~al.
\newblock {\em Convex analysis and monotone operator theory in Hilbert spaces},
  volume 408.
\newblock Springer, 2011.

\bibitem{cai2019once}
Han Cai, Chuang Gan, Tianzhe Wang, Zhekai Zhang, and Song Han.
\newblock Once-for-all: Train one network and specialize it for efficient
  deployment.
\newblock In {\em International Conference on Learning Representations}, 2019.

\bibitem{8897011}
H. {Cai}, J. {Lin}, Y. {Lin}, Z. {Liu}, K. {Wang}, T. {Wang}, L. {Zhu}, and S.
  {Han}.
\newblock Automl for architecting efficient and specialized neural networks.
\newblock {\em IEEE Micro}, 40(1):75--82, 2020.

\bibitem{Chang_2017_ICCV}
Feng-Ju Chang, Anh Tuan~Tran, Tal Hassner, Iacopo Masi, Ram Nevatia, and Gerard
  Medioni.
\newblock Faceposenet: Making a case for landmark-free face alignment.
\newblock In {\em Proceedings of the IEEE International Conference on Computer
  Vision (ICCV) Workshops}, Oct 2017.

\bibitem{Chen_2019_ICCV}
Chen Chen, Qifeng Chen, Minh~N. Do, and Vladlen Koltun.
\newblock Seeing motion in the dark.
\newblock In {\em Proceedings of the IEEE/CVF International Conference on
  Computer Vision (ICCV)}, October 2019.

\bibitem{chen2020orthant}
Tianyi Chen, Tianyu Ding, Bo Ji, Guanyi Wang, Jing Tian, Yixin Shi, Sheng Yi,
  Xiao Tu, and Zhihui Zhu.
\newblock Orthant based proximal stochastic gradient method for l1-regularized
  optimization.
\newblock In {\em Machine Learning and Knowledge Discovery in Databases}, pages
  57--73, Cham, 2021. Springer International Publishing.

\bibitem{chin2020towards}
Ting-Wu Chin, Ruizhou Ding, Cha Zhang, and Diana Marculescu.
\newblock Towards efficient model compression via learned global ranking.
\newblock In {\em Proceedings of the IEEE/CVF Conference on Computer Vision and
  Pattern Recognition}, pages 1518--1528, 2020.

\bibitem{deng2009imagenet}
Jia Deng, Wei Dong, Richard Socher, Li-Jia Li, Kai Li, and Li Fei-Fei.
\newblock Imagenet: A large-scale hierarchical image database.
\newblock In {\em 2009 IEEE conference on computer vision and pattern
  recognition}, pages 248--255. Ieee, 2009.

\bibitem{9043731}
L. {Deng}, G. {Li}, S. {Han}, L. {Shi}, and Y. {Xie}.
\newblock Model compression and hardware acceleration for neural networks: A
  comprehensive survey.
\newblock {\em Proceedings of the IEEE}, 108(4):485--532, 2020.

\bibitem{ding2020lossless}
Xiaohan Ding, Tianxiang Hao, Ji Liu, Jungong Han, Yuchen Guo, and Guiguang
  Ding.
\newblock Lossless cnn channel pruning via decoupling remembering and
  forgetting.
\newblock {\em arXiv preprint arXiv:2007.03260}, 2020.

\bibitem{Gorji_2018_CVPR}
Siavash Gorji and James~J. Clark.
\newblock Going from image to video saliency: Augmenting image salience with
  dynamic attentional push.
\newblock In {\em Proceedings of the IEEE Conference on Computer Vision and
  Pattern Recognition (CVPR)}, June 2018.

\bibitem{guan2019specnet}
Bochen Guan, Jinnian Zhang, William~A Sethares, Richard Kijowski, and Fang Liu.
\newblock Specnet: Spectral domain convolutional neural network.
\newblock {\em arXiv preprint arXiv:1905.10915}, 2019.

\bibitem{he2016deep}
Kaiming He, Xiangyu Zhang, Shaoqing Ren, and Jian Sun.
\newblock Deep residual learning for image recognition.
\newblock In {\em Proceedings of the IEEE conference on computer vision and
  pattern recognition}, pages 770--778, 2016.

\bibitem{he2018amc}
Yihui He, Ji Lin, Zhijian Liu, Hanrui Wang, Li-Jia Li, and Song Han.
\newblock Amc: Automl for model compression and acceleration on mobile devices.
\newblock In {\em Proceedings of the European Conference on Computer Vision
  (ECCV)}, pages 784--800, 2018.

\bibitem{he2019filter}
Yang He, Ping Liu, Ziwei Wang, Zhilan Hu, and Yi Yang.
\newblock Filter pruning via geometric median for deep convolutional neural
  networks acceleration.
\newblock In {\em Proceedings of the IEEE/CVF Conference on Computer Vision and
  Pattern Recognition}, pages 4340--4349, 2019.

\bibitem{he2017channel}
Yihui He, Xiangyu Zhang, and Jian Sun.
\newblock Channel pruning for accelerating very deep neural networks.
\newblock In {\em Proceedings of the IEEE International Conference on Computer
  Vision}, pages 1389--1397, 2017.

\bibitem{huang2017densely}
Gao Huang, Zhuang Liu, Laurens Van Der~Maaten, and Kilian~Q Weinberger.
\newblock Densely connected convolutional networks.
\newblock In {\em Proceedings of the IEEE conference on computer vision and
  pattern recognition}, pages 4700--4708, 2017.

\bibitem{9022101}
A. {Ignatov}, R. {Timofte}, A. {Kulik}, S. {Yang}, K. {Wang}, F. {Baum}, M.
  {Wu}, L. {Xu}, and L. {Van Gool}.
\newblock Ai benchmark: All about deep learning on smartphones in 2019.
\newblock In {\em 2019 IEEE/CVF International Conference on Computer Vision
  Workshop (ICCVW)}, pages 3617--3635, 2019.

\bibitem{kingma2015adam}
Diederik~P Kingma and Jimmy Ba.
\newblock Adam: A method for stochastic optimization.
\newblock In {\em ICLR}, 2015.

\bibitem{krizhevsky2009learning}
Alex Krizhevsky, Geoffrey Hinton, et~al.
\newblock Learning multiple layers of features from tiny images.
\newblock 2009.

\bibitem{kusupati2020soft}
Aditya Kusupati, Vivek Ramanujan, Raghav Somani, Mitchell Wortsman, Prateek
  Jain, Sham Kakade, and Ali Farhadi.
\newblock Soft threshold weight reparameterization for learnable sparsity.
\newblock In {\em International Conference on Machine Learning}, pages
  5544--5555. PMLR, 2020.

\bibitem{li2016pruning}
Hao Li, Asim Kadav, Igor Durdanovic, Hanan Samet, and Hans~Peter Graf.
\newblock Pruning filters for efficient convnets.
\newblock {\em International Conference on Learning Representations}, 2017.

\bibitem{ijcai2020-363}
Hang Li, Chen Ma, Wei Xu, and Xue Liu.
\newblock Feature statistics guided efficient filter pruning.
\newblock In Christian Bessiere, editor, {\em Proceedings of the Twenty-Ninth
  International Joint Conference on Artificial Intelligence, {IJCAI-20}}, pages
  2619--2625. International Joint Conferences on Artificial Intelligence
  Organization, 7 2020.
\newblock Main track.

\bibitem{Li_2019_CVPR}
Jiashi Li, Qi Qi, Jingyu Wang, Ce Ge, Yujian Li, Zhangzhang Yue, and Haifeng
  Sun.
\newblock Oicsr: Out-in-channel sparsity regularization for compact deep neural
  networks.
\newblock In {\em Proceedings of the IEEE/CVF Conference on Computer Vision and
  Pattern Recognition (CVPR)}, June 2019.

\bibitem{10.1007/978-3-030-58598-3_36}
Yawei Li, Shuhang Gu, Kai Zhang, Luc Van~Gool, and Radu Timofte.
\newblock Dhp: Differentiable meta pruning via hypernetworks.
\newblock In Andrea Vedaldi, Horst Bischof, Thomas Brox, and Jan-Michael Frahm,
  editors, {\em Computer Vision -- ECCV 2020}, pages 608--624, Cham, 2020.
  Springer International Publishing.

\bibitem{lin2020hrank}
Mingbao Lin, Rongrong Ji, Yan Wang, Yichen Zhang, Baochang Zhang, Yonghong
  Tian, and Ling Shao.
\newblock Hrank: Filter pruning using high-rank feature map.
\newblock In {\em Proceedings of the IEEE/CVF Conference on Computer Vision and
  Pattern Recognition}, pages 1529--1538, 2020.

\bibitem{lin2019towards}
Shaohui Lin, Rongrong Ji, Chenqian Yan, Baochang Zhang, Liujuan Cao, Qixiang
  Ye, Feiyue Huang, and David Doermann.
\newblock Towards optimal structured cnn pruning via generative adversarial
  learning.
\newblock In {\em Proceedings of the IEEE/CVF Conference on Computer Vision and
  Pattern Recognition}, pages 2790--2799, 2019.

\bibitem{liu2015sparse}
Baoyuan Liu, Min Wang, Hassan Foroosh, Marshall Tappen, and Marianna Pensky.
\newblock Sparse convolutional neural networks.
\newblock In {\em Proceedings of the IEEE conference on computer vision and
  pattern recognition}, pages 806--814, 2015.

\bibitem{liu2017learning}
Zhuang Liu, Jianguo Li, Zhiqiang Shen, Gao Huang, Shoumeng Yan, and Changshui
  Zhang.
\newblock Learning efficient convolutional networks through network slimming.
\newblock In {\em Proceedings of the IEEE International Conference on Computer
  Vision}, pages 2736--2744, 2017.

\bibitem{liu2018rethinking}
Zhuang Liu, Mingjie Sun, Tinghui Zhou, Gao Huang, and Trevor Darrell.
\newblock Rethinking the value of network pruning.
\newblock In {\em International Conference on Learning Representations}, 2018.

\bibitem{liu2018frequency}
Zhenhua Liu, Jizheng Xu, Xiulian Peng, and Ruiqin Xiong.
\newblock Frequency-domain dynamic pruning for convolutional neural networks.
\newblock In {\em Proceedings of the 32nd International Conference on Neural
  Information Processing Systems}, pages 1051--1061, 2018.

\bibitem{luo2017thinet}
Jian-Hao Luo, Jianxin Wu, and Weiyao Lin.
\newblock Thinet: A filter level pruning method for deep neural network
  compression.
\newblock In {\em Proceedings of the IEEE international conference on computer
  vision}, pages 5058--5066, 2017.

\bibitem{ma2020pconv}
Xiaolong Ma, Fu-Ming Guo, Wei Niu, Xue Lin, Jian Tang, Kaisheng Ma, Bin Ren,
  and Yanzhi Wang.
\newblock Pconv: The missing but desirable sparsity in dnn weight pruning for
  real-time execution on mobile devices.
\newblock In {\em Proceedings of the AAAI Conference on Artificial
  Intelligence}, volume~34, pages 5117--5124, 2020.

\bibitem{Meng2020pruning}
Fanxu Meng, Hao Cheng, Ke Li, Huixiang Luo, Xiaowei Guo, Guangming Lu, and Xing
  Sun.
\newblock Pruning filter in filter.
\newblock In {\em Advances in Neural Information Processing Systems},
  volume~33, pages 17629--17640, 2020.

\bibitem{niu2020patdnn}
Wei Niu, Xiaolong Ma, Sheng Lin, Shihao Wang, Xuehai Qian, Xue Lin, Yanzhi
  Wang, and Bin Ren.
\newblock Patdnn: Achieving real-time dnn execution on mobile devices with
  pattern-based weight pruning.
\newblock In {\em Proceedings of the Twenty-Fifth International Conference on
  Architectural Support for Programming Languages and Operating Systems}, pages
  907--922, 2020.

\bibitem{peng2018extreme}
Bo Peng, Wenming Tan, Zheyang Li, Shun Zhang, Di Xie, and Shiliang Pu.
\newblock Extreme network compression via filter group approximation.
\newblock In {\em Proceedings of the European Conference on Computer Vision
  (ECCV)}, pages 300--316, 2018.

\bibitem{raihan2020sparse}
Md~Aamir Raihan and Tor~M Aamodt.
\newblock Sparse weight activation training.
\newblock {\em arXiv preprint arXiv:2001.01969}, 2020.

\bibitem{real2019regularized}
Esteban Real, Alok Aggarwal, Yanping Huang, and Quoc~V Le.
\newblock Regularized evolution for image classifier architecture search.
\newblock In {\em Proceedings of the aaai conference on artificial
  intelligence}, volume~33, pages 4780--4789, 2019.

\bibitem{ronneberger2015u}
Olaf Ronneberger, Philipp Fischer, and Thomas Brox.
\newblock U-net: Convolutional networks for biomedical image segmentation.
\newblock In {\em International Conference on Medical image computing and
  computer-assisted intervention}, pages 234--241. Springer, 2015.

\bibitem{roth2008group}
Volker Roth and Bernd Fischer.
\newblock The group-lasso for generalized linear models: uniqueness of
  solutions and efficient algorithms.
\newblock In {\em Proceedings of the 25th international conference on Machine
  learning}, pages 848--855, 2008.

\bibitem{guo2020dmcp}
Guo Shaopeng, Wang Yujie, Li Quanquan, and Junjie Yan.
\newblock Dmcp: Differentiable markov channel pruning for neural networks.
\newblock In {\em IEEE Conference on Computer Vision and Pattern Recognition
  (CVPR)}, 2020.

\bibitem{simon2013sparse}
Noah Simon, Jerome Friedman, Trevor Hastie, and Robert Tibshirani.
\newblock A sparse-group lasso.
\newblock {\em Journal of computational and graphical statistics},
  22(2):231--245, 2013.

\bibitem{Simonyan15}
Karen Simonyan and Andrew Zisserman.
\newblock Very deep convolutional networks for large-scale image recognition.
\newblock In {\em International Conference on Learning Representations}, 2015.

\bibitem{sun2020learning}
Zhongkai Sun, Prathusha Sarma, William Sethares, and Yingyu Liang.
\newblock Learning relationships between text, audio, and video via deep
  canonical correlation for multimodal language analysis.
\newblock In {\em Proceedings of the AAAI Conference on Artificial
  Intelligence}, volume~34, pages 8992--8999, 2020.

\bibitem{szegedy2015going}
Christian Szegedy, Wei Liu, Yangqing Jia, Pierre Sermanet, Scott Reed, Dragomir
  Anguelov, Dumitru Erhan, Vincent Vanhoucke, and Andrew Rabinovich.
\newblock Going deeper with convolutions.
\newblock In {\em Proceedings of the IEEE conference on computer vision and
  pattern recognition}, pages 1--9, 2015.

\bibitem{tan2019efficientnet}
Mingxing Tan and Quoc Le.
\newblock Efficientnet: Rethinking model scaling for convolutional neural
  networks.
\newblock In {\em International Conference on Machine Learning}, pages
  6105--6114. PMLR, 2019.

\bibitem{wang2020deep}
Fangxin Wang, Miao Zhang, Xiangxiang Wang, Xiaoqiang Ma, and Jiangchuan Liu.
\newblock Deep learning for edge computing applications: A state-of-the-art
  survey.
\newblock {\em IEEE Access}, 8:58322--58336, 2020.

\bibitem{wang2017factorized}
Min Wang, Baoyuan Liu, and Hassan Foroosh.
\newblock Factorized convolutional neural networks.
\newblock In {\em Proceedings of the IEEE International Conference on Computer
  Vision Workshops}, pages 545--553, 2017.

\bibitem{wen2016learning}
Wei Wen, Chunpeng Wu, Yandan Wang, Yiran Chen, and Hai Li.
\newblock Learning structured sparsity in deep neural networks.
\newblock {\em arXiv preprint arXiv:1608.03665}, 2016.

\bibitem{zhao2019variational}
Chenglong Zhao, Bingbing Ni, Jian Zhang, Qiwei Zhao, Wenjun Zhang, and Qi Tian.
\newblock Variational convolutional neural network pruning.
\newblock In {\em Proceedings of the IEEE/CVF Conference on Computer Vision and
  Pattern Recognition}, pages 2780--2789, 2019.

\bibitem{zhu2017prune}
Michael Zhu and Suyog Gupta.
\newblock To prune, or not to prune: exploring the efficacy of pruning for
  model compression.
\newblock {\em arXiv preprint arXiv:1710.01878}, 2017.

\end{thebibliography}
}

\newpage
\appendix

\section{SMOF Pruning for U-Net}
\label{App: U-Net pruning procedures}
\textbf{SMOF pruning procedures for U-Net} First, we keep all the elements in Filter Masks to be $1$ ($\beta = 0$, $\delta_{\text{FM}} = 0$, and $r = 0$) and set $\alpha = 1e-6$, $\rho_{\text{FS}} = 0$ for $50$ epochs. Then, we either 1) keep $\alpha = 1e-6$ and set $\rho_{\text{FS}} = 0.30$, $\beta = 3e-4$, $\delta_{\text{FM}} = 0.02$, and $r = 1$ for another 50 epochs, or 2) set $\alpha = 2e-6$, and $\rho_{\text{FS}} = 0.35$, keep $\beta = 0$, $\delta_{\text{FM}} = 0$, and $r = 0$ for 50 epochs. We call the obtained models SMOF-1 and SMOF-2, respectively. Next, we stop pruning and fix their network structures by setting $\alpha = \rho_{\text{FS}} = \beta = \delta_{\text{FM}} = 0$, and apply  Adam optimizer \cite{kingma2015adam} with learning rate $= 1e-4$ with for 100 epochs.

\section{Structure of Pruned ResNet18}
\label{App: ResNet18 structure}

In this section, we provide the structure of the structure of a pruned ResNet18. Specifically, we provide a comparison of the original ResNet-18 (FLOPs = 1842.78M) with the SMOF-1 in Sec. \ref{subsec: ResNet18 test} (FLOPs = 1271.17M).

\begin{table}[H]
\begin{tabular}{c|cc}
                         & ResNet18 & SMOF-1 \\ \hline\hline
conv1                    & (64, 7)    & (64, 5)  \\ \hline
layer1.0.conv1           & (64, 3)  & (64, 1)  \\ \hline
layer1.0.conv2           & (64, 3)    & (64, 3) \\ \hline
layer1.1.conv1           & (64, 3)    & (64, 3)  \\ \hline
layer1.1.conv2           & (64, 3)    & (64, 3)  \\ \hline
layer2.0.conv1           & (128, 3)   & (128, 3) \\ \hline
layer2.0.conv2           & (128, 3)   & (128, 1) \\ \hline
layer2.0.downsample.conv & (128, 1)   & (128, 1) \\ \hline
layer2.1.conv1           & (128, 3)   & (128, 3) \\ \hline
layer2.1.conv2           & (128, 3)   & (128, 1) \\ \hline
layer3.0.conv1           & (256, 3)   & (256, 3) \\ \hline
layer3.0.conv2           & (256, 3)   & (256, 3) \\ \hline
layer3.0.downsample.conv & (256, 1)   & (256, 1) \\ \hline
layer3.1.conv1           & (256, 3)   & (256, 3) \\ \hline
layer3.1.conv2           & (256, 3)   & (256, 1) \\ \hline
layer4.0.conv1           & (512, 3)   & (512, 3) \\ \hline
layer4.0.conv2           & (512, 3)   & (512, 3) \\ \hline
layer4.0.downsample.conv & (512, 1)   & (512, 1) \\ \hline
layer4.1.conv1           & (512, 3)   & (512, 3) \\ \hline
layer4.1.conv2           & (512, 3)   & (512, 1) \\ \hline
\end{tabular}
\caption{\textbf{Comparison of (output channels, kernel size) of the Conv layers in ResNet18 and SMOF-1.} Note that SMOF only reduces the kernel sizes for ResNet18.}
\end{table}

\end{document}